# LRSVRG-IMC: An SVRG-Based Algorithm for Low-Rank Inductive Matrix Completion

Shangrong Yu[1[0000-0002-1219-4960]], Yuxin Chen [1[0000-0002-6274-5206]] and Hejun Wu [1*[0000-0001-9758-5698]]

[1] Sun Yat-sen University, Guangzhou 510006, China
wuhejun@mail.sysu.edu.cn

**Abstract.** Low-rank inductive matrix completion (IMC) is currently widely used in IoT data completion, recommendation systems, and so on, as the side information in IMC has demonstrated great potential in reducing sample complexity and computational complexity. Unfortunately, the saddle point remains a major obstacle for the convergence of the nonconvex solutions to IMC. What's more, carefully choosing the initial solution alone does not usually help remove the saddle points. To address this problem, we propose a stochastic variance reduction gradient-based algorithm called LRSVRG- IMC. LRSVRG-IMC can escape from the saddle points under various low-rank and sparse conditions with a properly chosen initial input. We also prove that LRSVRG-IMC achieves both a linear convergence rate and a near-optimal sample complexity. The superiority and applicability of LRSVRG-IMC are verified via experiments on synthetic datasets.

## 1 Introduction

Matrix completion is a powerful tool for signal processing and has been widely used in IoT data completion, recommendation systems and image processing applications. In practice, side information, also named feature, can be directly used for matrix completion in addition to the observed entries. For instance, IMC (inductive matrix completion) uses side information to recover the unknown matrix (Jain and Dhillon, 2013). Xu et al. (2010) designed a method with convex relaxation that requires only $O(rn\log n\log d)$ samples to recover the true matrix. The disadvantage of their method is the long latency, as it needs to solve a nuclear norm minimization problem involving a non-strongly convex objective function. The alternating minimization method by Jain and Dhillon (2013) achieves a locally linear convergence rate with $O(r^3n^2\log n\log\frac{1}{\epsilon})$ ($\epsilon$ is the convergence precision) samples. Nevertheless, the convergence rate is dependent on $\frac{1}{\epsilon}$, which indicates that exact recovery is impossible.

In contrast, nonconvex approaches achieved better overall performance (Xu et al., 2013; Shin et al., 2015; Chiang et al., 2015; Soni et al., 2017; Huang et al., 2017; Zhang et al., 2018b). GD-IMC (Zhang et al. 2018b)is a typical efficient nonconvex solution that needs only $O(r^2 n \log n \log d)$ samples while achieving a linear rate of convergence. However, we observed that such an approach demonstrated unstable performance under different levels of sparsity. When the observed matrix becomes sparse, GD-IMC tended to be trapped in a saddle point. This observation is illustrated by figures in the experimental section. Significant instability is not observed in the approaches such as Xu et al. (2013); Shin et al. (2015); Chiang et al. (2015); Soni et al. (2017); Huang et al. (2017). Nevertheless, none of them achieves a sample complexity as low and a convergence rate as high as GD-IMC.

In this paper, inspired by Wang et al. (2017), we propose a new SVRG (stochastic variance reduction gradient)-based approach, called LRSVRG-IMC, to solve the IMC problem while trying to escape from saddle points. Our contributions contain: (1) We design LRSVRG-IMC algorithm, whose computational and sample complexity are $O(r^2 n \log n \log d \log\frac{1}{\epsilon} + r^3 n^2 d^{3/2} \log d \log\frac{1}{\epsilon})$ and $O(r^2 n \log n \log d)$, respectively. (2) We found that with the initial input carefully chosen to be close to the optimum, the projection step can be omitted, which reduces the computational complexity to $O\left(r^2 n \log n \log d \log\frac{1}{\epsilon}\right)$. (3)The experimental results demonstrate that LRSVRG-IMC can escape from almost all saddle points in the test datasets and reach an even lower sample complexity of $O(rn \log n)$ in practice, which is quite close to the theoretical bound with an extra logarithm factor.

# 2 SVRG-based Algorithm

## 2.1 Notations

We use upper case letters such as $\mathbf{A}$ to represent matrices and lower case letters with square brackets to represent sets, e.g., $[d]$ is used to denote the index set $\{1, 2, \ldots, d\}$. Likewise, $d_1 \times d_2$ stands for $\{(i,j) | i \in [d_1], j \in [d_2]\}$. Specifically, a capital with a star, such as $\mathbf{A}^*$, is used for the optimum of a certain optimization problem. Let $\mathbf{A}_{i,*}$, $\mathbf{A}_{*,j}$, $A_{ij}$ be the $i$-th row, $j$-th column and $ij$-th entry of matrix $\mathbf{A}$, respectively. The $k$-th largest singular value of matrix $\mathbf{A}$ is denoted as $\sigma_k(\mathbf{A})$, and the projection of matrix $\mathbf{A}$ onto the index set $\Omega$ is denoted by $P_\Omega(\mathbf{A})$, where $P_\Omega(\mathbf{A})$ equals $A_{ij}$ if $(i, j) \in \Omega$ and zero otherwise. For the sake of simplicity, we also denote $d = \max\{d_1, d_2\}$ and $n = \max\{n_1, n_2\}$. For any two sequences $\{x_n\}$ and $\{y_n\}$, if there exists a positive constant $c$ so that $x_n \leq cy_n$, we say $x_n = O(y_n)$.

## 2.2 Problem Setup

Now, we formulate the problem for LRSVRG-IMC. Let $\mathbf{L} \in \mathbb{R}^{d_1 \times d_2}$ be the observed matrix with missing values and $\mathbf{X}_L \in \mathbb{R}^{d_1 \times n_1}$ and $\mathbf{X}_R \in \mathbb{R}^{d_2 \times n_2}$ be the feature matrices. IMC aims at learning a relatively low-dimension low-rank matrix $\mathbf{M}^* \in \mathbb{R}^{n_1 \times n_2}$, where $\mathbf{L}^* = \mathbf{X}_L \mathbf{M}^* \mathbf{X}_R^{\mathrm{T}}$, to obtain the unknown higher-dimension low-rank matrix $\mathbf{L}^* \in$ $\mathbb{R}^{d_1 \times d_2}$ ($d_1 \geq n_1 \geq r$, $d_2 \geq n_2 \geq r$). Here, entries of $\mathbf{L}^*$ are independently observed with probability $p_{ij} \in (0, 1)$. Specifically, we consider the Bernoulli model for sampling:

$$L_{ij} = \begin{cases} L_{ij}^*, & \text{with probability } p; \\ *, & \text{else.} \end{cases} \tag{2}$$

Let $\Omega$ be the index set such that $\Omega = \{(i, j) \in [d_1] \times [d_2] | L_{ij} \neq *\}$. After the projection step, we have $P_\Omega(\mathbf{L}) = P_\Omega(\mathbf{L}^*)$.

Recall that $\mathbf{M}^*$ is of rank $r$. Given SVD $\mathbf{M}^* = \bar{\mathbf{U}}^* \boldsymbol{\Sigma}^* \bar{\mathbf{V}}^{*T}$, there are $r$ nonzero entries in the diagonal matrix $\boldsymbol{\Sigma}^*$. Denote $\sigma_r{}^*(\mathbf{M}^*) = \sigma_r{}^*$, and these $r$ entries, also known as singular values, can be expressed as $\sigma_1{}^* \geq \sigma_2{}^* \geq \cdots \geq \sigma_r{}^* > 0$ from left to right, with the condition number $\kappa = \frac{\sigma_1^*}{\sigma_r^*}$.

To fully exploit the side information, we also impose the standard feasibility condition (SFC) (Xu et al., 2013;) for the IMC model: $col(\mathbf{L}^*) \subseteq col(\mathbf{X}_L)$ and $col(\mathbf{L}^{*T}) \subseteq col(\mathbf{X}_R)$, which means that the column space of the unknown matrix is contained in the column space of the feature matrices. In this way, the low-rankness of the unknown matrix is expressed in the feature matrices for better performance.

Note that $\mathbf{L}^* = \mathbf{X}_L \mathbf{M}^* \mathbf{X}_R^{\mathrm{T}}$ in IMC. In this paper, we assume that both $\mathbf{X}_L$ and $\mathbf{X}_R$ are orthogonal matrices, i.e., $\mathbf{X}_L^{\mathrm{T}} \mathbf{X}_L = \mathbf{I}_{n_1}$, $\mathbf{X}_R^{\mathrm{T}} \mathbf{X}_R = \mathbf{I}_{n_2}$. In addition, recovering a matrix with too few observed entries is unlikely (Gross, 2011). Therefore, we impose $\mu$-incoherence for the unknown matrix $\mathbf{L}^*$, as shown in (Zhang et al., 2018b). Due to the property of the SFC, we also impose a similar constraint for the two feature matrices, as shown in Property 2 (Zhang et al., 2018b).

*Property 1($\mu_1$-incoherence).* The feature matrices $\mathbf{X}_L$ and $\mathbf{X}_R$ satisfy $\mu_1$-incoherence, i.e., $\|\mathbf{X}_L\|_{2,\infty} \leq \sqrt{\frac{\mu_1 n_1}{d_1}}$ and $\|\mathbf{X}_R\|_{2,\infty} \leq \sqrt{\frac{\mu_1 n_2}{d_2}}$.

With these properties, the inductive matrix completion problem can now be formulated as Eq. (3),

$$\begin{gathered} \min_{\mathbf{M} \in \mathbb{R}^{n_1 \times n_2}} \frac{1}{2p} \|P_\Omega(\mathbf{X}_L \mathbf{M} \mathbf{X}_R^{\mathrm{T}} - \mathbf{L})\|_F^2, \\ s.t.\ rank(\mathbf{M}) \leq r, \end{gathered} \tag{3}$$

Where $p = \frac{|\Omega|}{d_1 d_2}$ is the sample rate for the observed matrix.

In our proposed approach, we take two steps further to reduce the complexity in Eq. (3). First, we make use of Burer-Monteiro decomposition to guarantee the low-rankness of the solution. Second, we introduce an extra penalty term to balance the two submatrices obtained from the decomposition. In this paper, we adopt the following model:

$$\min_{\substack{\mathbf{U} \in \mathbb{R}^{n_1 \times r} \\ \mathbf{V} \in \mathbb{R}^{n_2 \times r}}} F_\Omega(\mathbf{U}, \mathbf{V}) = \frac{1}{2p} \|P_\Omega(\mathbf{X}_L \mathbf{U} \mathbf{V}^{\mathrm{T}} \mathbf{X}_R^{\mathrm{T}} - \mathbf{L}^*)\|_F^2 + \mathcal{R}(\mathbf{U}, \mathbf{V}), \tag{4}$$

Here, we adopt the penalty term as: $\mathcal{R}(\mathbf{U}, \mathbf{V}) = \frac{1}{8} \|\mathbf{U}^{\mathrm{T}} \mathbf{U} - \mathbf{V}^{\mathrm{T}} \mathbf{V}\|_F^2$. For convenience, we denote: $\mathcal{L}_\Omega(\mathbf{X}_L \mathbf{U} \mathbf{V}^{\mathrm{T}} \mathbf{X}_R^{\mathrm{T}}) = \frac{1}{2p} \|P_\Omega(\mathbf{X}_L \mathbf{U} \mathbf{V}^{\mathrm{T}} \mathbf{X}_R^{\mathrm{T}} - \mathbf{L}^*)\|_F^2$. Then, Eq. (4) can be simplified as$\min_{\mathbf{U},\mathbf{V}} F_\Omega(\mathbf{U}, \mathbf{V}) = \mathcal{L}_\Omega(\mathbf{X}_L \mathbf{U} \mathbf{V}^{\mathrm{T}} \mathbf{X}_R^{\mathrm{T}}) + \mathcal{R}(\mathbf{U}, \mathbf{V})$.

**Algorithm 1** Initialization

**Input:** observed matrix $P_\Omega(\mathbf{L}^*)$, feature matrices $\mathbf{X}_L$ and $\mathbf{X}_R$, observed index set $\Omega$, number of subsets $n$.
1: Randomly split index set $\Omega$ into $n$ subsets, namely, $\Omega_1, \Omega_2, ..., \Omega_n$, where the size of each is $\frac{|\Omega|}{n}$
2: Randomly choose $\frac{n}{2}$ subsets to form subset $\Omega_0$, the size of which is thus $\frac{|\Omega|}{2}$
3: Perform SVD-$r$ based on $\Omega_0$: $[\widetilde{\mathbf{U}}_0, \boldsymbol{\Sigma}_0, \widetilde{\mathbf{V}}_0] = \mathrm{SVD}_r({p_0}^{-1} P_{\Omega_0}(\mathbf{L}^*))$
4: $\widetilde{\mathbf{U}}^0 = \mathbf{X}_L^{\mathrm{T}} \widetilde{\mathbf{U}}_0 \boldsymbol{\Sigma}_0^{\frac{1}{2}}$, $\widetilde{\mathbf{V}}^0 = \mathbf{X}_R^{\mathrm{T}} \widetilde{\mathbf{V}}_0 \boldsymbol{\Sigma}_0^{\frac{1}{2}}$
5: $\widetilde{\mathbf{Z}}_0 = [\widetilde{\mathbf{U}}^0; \widetilde{\mathbf{V}}^0]$
**Output:** $(\widetilde{\mathbf{U}}^0, \widetilde{\mathbf{V}}^0)$

To our knowledge, SVRG and Nesterov's momentum may have been the two most popular techniques to cope with saddle points thus far. Taking the structure of IMC into consideration, we consider that SVRG should be a more befitting approach. Recall that Eq. (4) is also a finite-sum problem. Hence, we first randomly split the index set $\boldsymbol{\Omega}$ into $n$ subsets, i.e., $\Omega_t$ $(t \in [n])$. In this way, the finite-sum format of Eq. (4) is obtained as: $F_\Omega(\mathbf{U}, \mathbf{V}) = \frac{1}{n} \sum_{t=1}^{n} F_{\Omega_t}(\mathbf{U}, \mathbf{V})$, where $F_{\Omega_t}(\mathbf{U}, \mathbf{V}) = \mathcal{L}_{\Omega_t}(\mathbf{X}_L \mathbf{U} \mathbf{V}^{\mathrm{T}} \mathbf{X}_R^{\mathrm{T}}) + \mathcal{R}(\mathbf{U}, \mathbf{V})$. Similar to the denotation $\mathcal{L}_\Omega(\mathbf{X}_L \mathbf{U} \mathbf{V}^{\mathrm{T}} \mathbf{X}_R^{\mathrm{T}})$, we have

$$\mathcal{L}_{\Omega_t}(\mathbf{X}_L \mathbf{U} \mathbf{V}^{\mathrm{T}} \mathbf{X}_R^{\mathrm{T}}) = \frac{1}{2p_t} \| P_{\Omega_t}(\mathbf{X}_L \mathbf{U} \mathbf{V}^{\mathrm{T}} \mathbf{X}_R^{\mathrm{T}} - \mathbf{L}^*) \|_F^2, \quad (7)$$

where $p_t = \frac{|\Omega_t|}{d_1 d_2}$ and $|\Omega_t| = \frac{|\Omega|}{n}$.

We need some extra definitions to clearly illustrate our theory. Let $\mathbf{Z} = [\mathbf{U}; \mathbf{V}]$ and $\mathbf{Z}^* = [\mathbf{U}^*; \mathbf{V}^*]$. We have three definitions as follows.

**Definition 1.** The distance between any two matrices with the same dimensions is defined as $\mathrm{d}(\mathbf{Z}, \mathbf{Z}^*) = \min_{\mathbf{R}} \|\mathbf{Z} - \mathbf{Z}^* \mathbf{R}\|_{\mathrm{F}}$, where $\mathbf{R}$ stands for an $r \times r$ transposed matrix.

**Definition 2.** The radius-$r$ neighborhood of $\mathbf{Z}^*$ is defined as: $\mathbb{B}(r) = \{\mathbf{Z} \in \mathbb{R}^{d_1 \times d_2} | d(\mathbf{Z}, \mathbf{Z}^*) \le r\}$.

**Definition 3.** Let $\mathbf{H}$ be the matrix form of the distance between any two matrices with the same dimensions as: $\mathbf{H} = \mathbf{Z} - \mathbf{Z}^* \widehat{\mathbf{R}}$, where $\widehat{\mathbf{R}} = \arg\min_{\mathbf{R} \in \mathbb{Q}_r} \|\mathbf{Z} - \mathbf{Z}^* \mathbf{R}\|_F$.

## 3 LRSVRG-IMC Algorithm

We propose Algorithm 2 to solve the IMC problem Eq. (4). For the sake of simplicity, $F_{\Omega_t}$ and $\mathcal{L}_{\Omega_t}$ are simplified as $F_t$ and $\mathcal{L}_t$ respectively. In Algorithm 2, the index set $\Omega$ is split into $n$ subsets to correspond to the finite-sum problem. In the variance reduction part, note that the stochastic terms here are slightly different from the standard ones.

In Algorithm 2, we take the alternating gradient descent on the two submatrices $\mathbf{U}$ and $\mathbf{V}$. This decomposition introduces more challenges, but the complexity can be sharply reduced (Jain and Dhillon, 2013; Zhang et al., 2018b; Wang et al., 2017). We also make slight changes to the submatrices: we use $\mathbf{V}^k$ to replace $\widetilde{\mathbf{V}}^k$ and $\mathbf{U}^k$ to replace $\widetilde{\mathbf{U}}^k$. The standard stochastic gradients are as follows:
$\nabla_{\mathbf{U}} F_{t_k}(\mathbf{U}^k, \mathbf{V}^k) - \mathbf{X}_L^{\mathrm{T}} \nabla \mathcal{L}_{t_k}(\tilde{\mathbf{L}}) \mathbf{X}_R \widetilde{\mathbf{V}}^k + \mathbf{X}_L^{\mathrm{T}} \nabla \mathcal{L}_\Omega(\tilde{\mathbf{L}}) \mathbf{X}_R \widetilde{\mathbf{V}}^k$ with regard to matrix $\mathbf{U}$ and $\nabla_{\mathbf{V}} F_{t_k}(\mathbf{U}^k, \mathbf{V}^k) - \mathbf{X}_R^{\mathrm{T}} \nabla \mathcal{L}_{t_k}(\tilde{\mathbf{L}})^{\mathrm{T}} \mathbf{X}_L \widetilde{\mathbf{U}}^k + \mathbf{X}_R^{\mathrm{T}} \nabla \mathcal{L}_\Omega(\tilde{\mathbf{L}})^{\mathrm{T}} \mathbf{X}_L \widetilde{\mathbf{U}}^k$ with regard to matrix $\mathbf{V}$. Our adaptation results in the so-called semi-stochastic term, which allows the whole gradi-

**Algorithm 2** LRSVRG-IMC

**Input:** observed matrix $P_\Omega(\mathbf{L}^*)$; feature matrices $\mathbf{X}_L$ and $\mathbf{X}_R$; step size $\tau$ ; number of iterations $S$ and $m$; loss function $F(\mathbf{U}, \mathbf{V})$; observed index set $\Omega$ and its subsets $\Omega_1, \Omega_2, \ldots, \Omega_n$, with $|\Omega_t| = |\Omega|/n, \forall t \in [n]$; initial solution $(\widetilde{\mathbf{U}}^0, \widetilde{\mathbf{V}}^0)$.

**for** $s = 1, 2, \ldots, S$ **do**
    $\widetilde{\mathbf{U}} = \widetilde{\mathbf{U}}^{s-1}, \widetilde{\mathbf{V}} = \widetilde{\mathbf{V}}^{s-1}, \tilde{\mathbf{L}} = \mathbf{X}_L \widetilde{\mathbf{U}} \widetilde{\mathbf{V}}^{\mathrm{T}} \mathbf{X}_R^{\mathrm{T}}$
    $\mathbf{U}^0 = \widetilde{\mathbf{U}}, \mathbf{V}^0 = \widetilde{\mathbf{V}}$
    **for** $k = 0, 1, \ldots, m$-1 **do**
        Randomly pick $t_k \in [n]$
        $\mathbf{U}^{k+1} = P_{C_1}\left(\mathbf{U}^k - \tau(\nabla_{\mathbf{U}} F_{t_k}(\mathbf{U}^k, \mathbf{V}^k) - \mathbf{X}_L^{\mathrm{T}} \nabla \mathcal{L}_{t_k}(\tilde{\mathbf{L}}) \mathbf{X}_R \mathbf{V}^k + \mathbf{X}_L^{\mathrm{T}} \nabla \mathcal{L}_\Omega(\tilde{\mathbf{L}}) \mathbf{X}_R \mathbf{V}^k)\right)$
        $\mathbf{V}^{k+1} = P_{C_2}\left(\mathbf{V}^k - \tau(\nabla_{\mathbf{V}} F_{t_k}(\mathbf{U}^k, \mathbf{V}^k) - \mathbf{X}_R^{\mathrm{T}} \nabla \mathcal{L}_{t_k}(\tilde{\mathbf{L}})^{\mathrm{T}} \mathbf{X}_L \mathbf{U}^k + \mathbf{X}_R^{\mathrm{T}} \nabla \mathcal{L}_\Omega(\tilde{\mathbf{L}})^{\mathrm{T}} \mathbf{X}_L \mathbf{U}^k)\right)$
    **end for**
    $\left(\widetilde{\mathbf{U}}^S, \widetilde{\mathbf{V}}^S\right) = \left(\mathbf{U}^{k'}, \mathbf{V}^{k'}\right)$, random $k' \in [m-1]$
**end for**

**Output:** $(\widetilde{\mathbf{U}}^S, \widetilde{\mathbf{V}}^S)$

ent to accumulate randomness through iterations. Recall that the randomness of SVRG is the key to escaping from a saddle point. Therefore, compared with standard forms, our semi-stochastic terms possess a greater ability to cope with a saddle point. A rigorous proof for the convergence under this adaptation is provided in the next section.

The details in defining the projection step $P_{C_i}(\mathbf{A})(i \in \{1, 2\})$ are presented as follows. For general SVRG algorithms (Reddi et al., 2016), convergence is almost guaranteed for a wide range of optimization problems. However, normally, the specific structure of the objective function is not taken into consideration, which implies that the standard SVRG may not satisfy properties 1 and 2 in our case.
To address this issue, we define a second type of projection onto the constrained set Ci so that the input for each iteration is well-structured, where the constrained sets have the form $\mathrm{C}_1 = \left\{ \mathrm{U} \in \mathbb{R}^{n_1 \times r} | \|\mathrm{X_L U}\|_{2,\infty} \leq \sqrt{\frac{\mu_0 r \sigma_1}{d_1}}\right.$, and $\mathrm{C}_2 = \left\{ \mathrm{V} \in \mathbb{R}^{n_2 \times r} | \|\mathrm{X_R V}\|_{2,\infty} \leq \sqrt{\frac{\mu_0 r \sigma_1}{d_2}}\right.$.

In practice, an interior point method or the Lagrangian method may be sufficient to achieve the projection. Since $\sigma_1$ is unknown here, we could use the singular value of the initial input as a rough approximation, while parameter $\mu_0$ could be tuned via traversal or according to practical demand.

Note that the input of Algorithm 2 must be close enough to the optimum in our analysis. Therefore, we adopt the first phase of Zhang et al. (2018b) as our generator for the initial solution, where we slightly modify the splitting pattern of the index set without any impact on its underlying logic. Algorithm 1 shows how the initialization process works. As is proven in Zhang et al. (2018b), Algorithm 1 guarantees that the output is within a small enough neighborhood of the optimum with a nearly optimal amount of samples. Surprisingly, in our experiments, we observed that the projection was not used at all in Algorithm 2 if the initial input is obtained from Algorithm 1. This implies that the computational complexity of our algorithm could be further reduced if the initial input is properly chosen. Additionally, the core theorem that illustrates how LRSVRG-IMC could converge to the optimum is placed in the appendix.

## 4 Evaluation

In this section, we compare two algorithms with our algorithm, i.e., AM-IMC and GD-

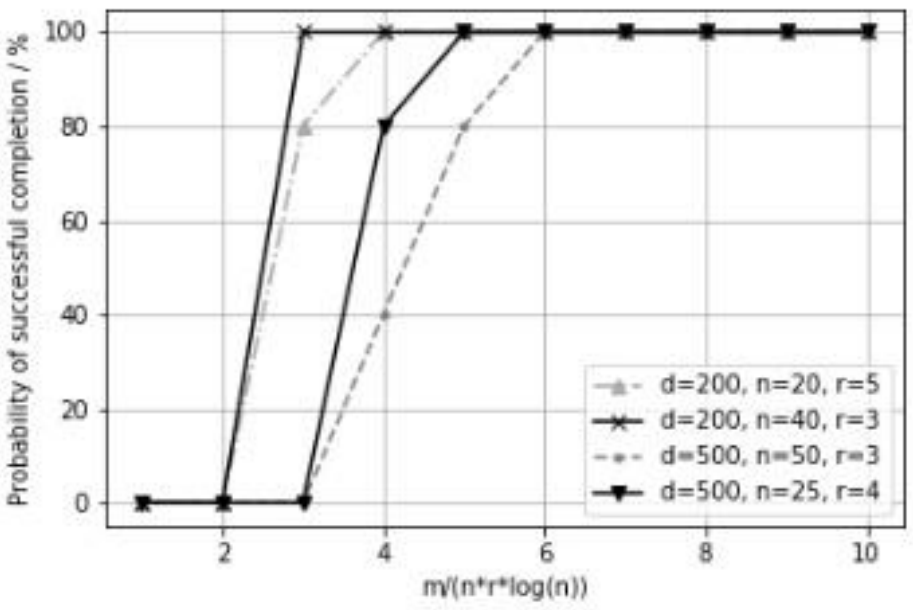

**Fig. 1.** Estimation of sample complexity.

IMC, on synthetic datasets to verify the superiority of LRSVRG-IMC.

### Synthetic Datasets

We first estimate the real sample complexity for four datasets. Their *d*, *n* and *r* are set as 200, 20, 5; 200, 40, 3; 500, 50, 3 and 500, 25, 4. We bound the number of outer iterations by 2000 times and generate 10 random test sets for each dataset. If the relative error satisfies $\frac{\|\mathbf{X}_L \mathbf{U}^S \mathbf{V}^{S^T} \mathbf{X}_R^T - \mathbf{L}^*\|_F}{\|\mathbf{L}^*\|_F} < 10^{-3}$, then we consider it a successful convergence. The results are plotted in Fig. 2, where the *x*-axis stands for the number of samples (note the denotation m here in the figure is distinguished from the random number *m*), and the *y*-axis stands for the probability of successful convergence on 10 test sets. For example, 80% implies that LRSVRG-IMC converged on 8 test sets. It is obvious that Algorithm 2 almost guarantees convergence at approximately $\frac{m}{nr\, log\, n} = 5$, which indicates that in practice, our sample complexity is likely to be much lower than the theoretical result.

We test two groups of sample rates, and 10 test sets are generated for each group. The first group is $p \in \{20\%, 30\%, 40\%\}$, on AM-IMC, GD-IMC and LRSVRG- IMC. The second one is $p \in \{3\%, 6\%, 9\%\}$, on GD-IMC and LRSVRG-IMC only, as AM-IMC does not work with too few samples. For experiments on the two groups, all parameters are tuned by K-folds, and all 10 test sets are run with the same group of well-tuned parameters.

The results of group one are displayed in Fig. 2, where the *x*-axis stands for an effective pass, i.e., a traversal on $|\mathbf{\Omega}|$ observed entries, and the *y*-axis stands for the relative error rate in the form of a logarithm with base 2. For the first ten effective passes, the value of y is slightly large; thus, we ignore them for a more detailed comparison. At first glance, there are no large differences with regard to the convergence rate in this group. In the earlier stage, LRSVRG-IMC even tends to converge more slowly, but it gradually speeds up, while the other two algorithms slow down.

The first group has relatively low sparsity, i.e., a high sample rate. Therefore, to further exploit the effectiveness of our work, we test higher sparsity data. The results of group 2 are displayed in Fig. 3. We also tried to run AM-IMC on this group of test sets, but it simply did not work. We can see that GD-IMC also poses poor convergence performance here, seemingly trapped in a certain saddle point far from the optimum. In contrast, our proposed algorithm keeps randomly attempting to escape and makes it quite soon.

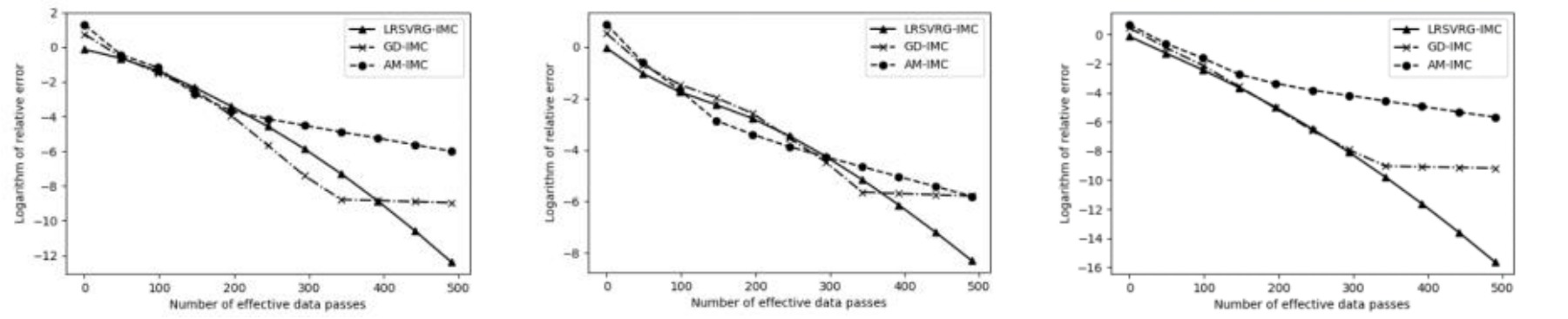

**Fig. 2.** Convergence under relatively high sample rates.

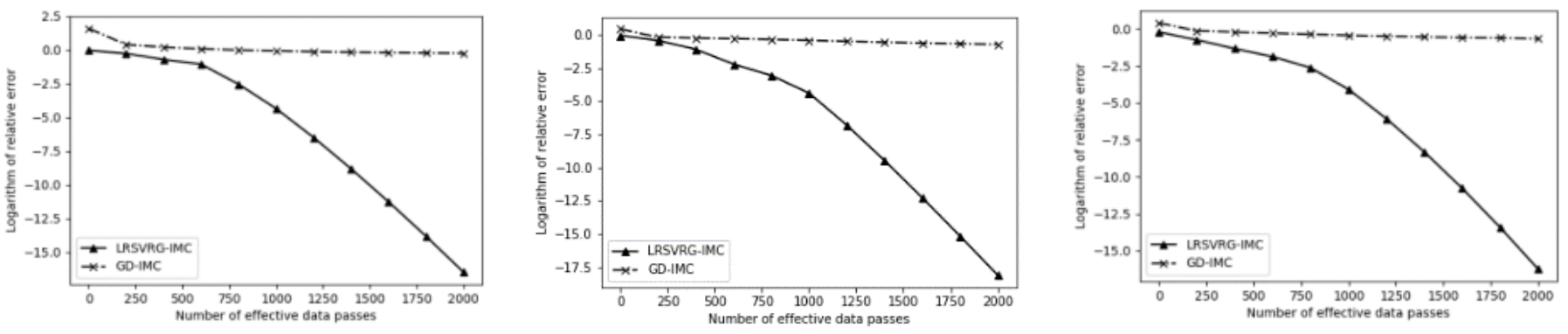

**Fig. 3.** Estimation of sample complexity.

In summary, our algorithm achieves a near-optimal sample complexity, fairly tight up to a logarithmic factor. Regarding the computational complexity, our algorithm requires fewer computations to reach the same convergence precision. In the cases of intractable saddle points, our method escapes with a much stronger ability compared with the state-of-the-art. This result is consistent with the theoretical analysis in the former sections.

# 5 Conclusions

In this paper, we propose the first SVRG-based algorithm for the inductive matrix completion problem, which achieves a near-optimal sample complexity and a linear rate of convergence. Our method guarantees convergence even when the observed matrix is of high sparsity. Moreover, the projection step may be omitted if the initial solution is carefully chosen. This can further improve the performance.

The appendix, named Appendix of IRSVRG-IMC, is uploaded to arXiv according to Reference 16.

# Appendix of LRSVRG-IMC

## Analysis of LRSVRG-IMC

*Theorem 1 (LRSVRG-IMC).* Suppose $\mathbf{L}^*$ satisfies properties 1 and 2. Then, there exist constant $\{c_i\}_{i=1}^{6}$, step size $\tau \le \min\{\frac{c_1}{\sigma_1(4r+1)}, \frac{c_2}{r\sigma_r}\}$, and random number $m = c_3 r$ such that as long as the number of samples satisfies $|\Omega| \ge c_4 \max\{\mu_1 n, \mu_0 r\kappa\}\, \mu_0 r^2 \kappa^2 \log n \log d$, and the initial input satisfies $\tilde{\mathbf{Z}}^0 \in \mathbb{B}(c_5\sqrt{\sigma_r})$, then, with a probability of at least $1-\frac{c_6}{d}$, the output of Algorithm 2 satisfies

$$E[d^2(\tilde{\mathbf{Z}}^S, \mathbf{Z}^*)] \le \rho^S E[d^2(\tilde{\mathbf{Z}}^0, \mathbf{Z}^*)],$$

Where $\rho \in (0,1)$.

Theorem 1 indicates that the sample complexity of Algorithm 2 is $O(r^2 n \log n \log d)$, which is linear with regard to parameter $n$, while that of LRMC is linear with regard to parameter $d$ (usually, we have $d\ n$). In addition, Algorithm 2 converges linearly to the optimal solution. Compared with AM-IMC, our sample complexity is reduced by at least $O(rn)$ times; thus, our approach is capable of coping with a sparser observed matrix. Compared with GD-IMC, our algorithm far exceeds GD-IMC in the cases of higher sparsity

To illustrate the overall performance of our algorithm, including initialization, we need to know how Algorithm 1 works. We put this result in Theorem 2. We refer readers to Zhang et al. (2018b) for detailed proofs of the theorem.

*Theorem 2 (Initialization).* Suppose the observed matrix $\mathbf{L}$ satisfies a Bernoulli model Eq. (2) and $\mathbf{L}^*$ satisfies properties 1 and 2. Then, $\forall\gamma \in (0,1)$, there exist constants $c_1$ and $c_2$ such that, as long as the number of samples satisfies $|\Omega| \ge \frac{c_1\mu_0\mu_1 r^2\kappa^2 n log\ d}{\gamma^2}$, with a probability of at least $1-\frac{c_2}{d}$, the output of Algorithm 1 satisfies $d(\tilde{\mathbf{Z}}^0, \mathbf{Z}^*) \le 4\gamma\sqrt{\sigma_r^*}$.

Theorem 2 shows that Algorithm 1 not only guarantees that the output is within a small enough region of the optimum but also achieves the same sample complexity as Algorithm 2 does.

In summary, with a properly chosen initializer, our proposed algorithm enjoys a linear rate of convergence as well as a (near) optimal sample complexity that nonconvex methods have reached thus far. This is the first algorithm that successfully introduces SVRG to the IMC model, reducing both the sample and computational complexities.

*Remark 1 (Computational complexity).* By Zhang et al. (2018b), the computational complexity of Algorithm 1 is $O(r|\Omega|)$. For Algorithm 2, the number of outer iterations is supposed to be $S = O(\log\frac{1}{\epsilon})$ to achieve $\rho^S E[d^2(\tilde{\mathbf{Z}}^0, \mathbf{Z}^*)] \le \epsilon$. For each outer iteration, a full gradient and $m$ stochastic gradients are computed, where $m = O(r)$. Thus, if we use the interior point method for the projection step, whose computational

complexity could be low as $O\left(r^2 n^2\, d^{\frac{3}{2}} log\, d\right)$, then the overall computational complexity is $O\left(\left(|\Omega| + r\left(|\Omega_t| + r^2 n^2\, d^{\frac{3}{2}} log\, d\right)\right) \log\frac{1}{\epsilon}\right)$.

Recall that $|\Omega| = O(r^2 n \log n \log d)$, so the formula above can be further expanded as $O\left(\left(1+\frac{r}{n}\right) r^2 n \log n \log d \log\frac{1}{\epsilon} + r^3 n^2\, d^{\frac{3}{2}} log\, d \log\frac{1}{\epsilon}\right)$. Note that for a low-rank matrix, we normally choose $r < n$ so that $1 + \frac{r}{n} < 2$. Considering that the projection step is skipped with the aid of Algorithm 1, the overall computational complexity of algorithms 1 and 2 is $O\left(r^3 n \log n \log d + r^2 n \log n\, log\, d \log\frac{1}{\epsilon}\right)$. As a comparison, to reach the same convergence precision, the state-of-the-art GD-IMC algorithm requires $O\left(r^3 n^2 d^{\frac{3}{2}} \log n \log d + r^3 n^2 \log n\, log\, d \log\frac{1}{\epsilon}\right)$ computations, which is at least $O(n)$ times higher than ours.

## A Detailed proof material of Theorem 1

In this appendix, we provide a complete and rigorous proof for Theorem 1. Recall that the objective function we aim at minimizing is

$$\min_{\substack{\mathbf{U}\in \mathbb{R}^{n_1 \times r} \\ \mathbf{V}\in \mathbb{R}^{n_2 \times r}}} F_\Omega(\mathbf{U},\mathbf{V}) = \frac{1}{2p}\|P_\Omega(\mathbf{X}_L\mathbf{U}\mathbf{V}^{\mathrm{T}}\mathbf{X}_R^{\mathrm{T}} - \mathbf{L}^*)\|_F^2 + \frac{1}{8}\|\mathbf{U}^{\mathbf{T}}\mathbf{U} - \mathbf{V}^{\mathbf{T}}\mathbf{V}\|_F^2,$$

which is equivalent to

$$\min_{\substack{\mathbf{U}\in \mathbb{R}^{n_1 \times r} \\ \mathbf{V}\in \mathbb{R}^{n_2 \times r}}} F_\Omega(\mathbf{U},\mathbf{V}) = \mathcal{L}\Omega(\mathbf{X}_L\mathbf{U}\mathbf{V}^{\mathrm{T}}\mathbf{X}_R^{\mathrm{T}}) + \frac{1}{8}\|\mathbf{U}^{\mathbf{T}}\mathbf{U} - \mathbf{V}^{\mathbf{T}}\mathbf{V}\|_F^2.$$

Thus, we obtain its gradients as

$$\nabla_{\mathbf{U}} F_\Omega(\mathbf{U},\mathbf{V}) = \nabla_{\mathbf{U}}\mathcal{L}_\Omega + \frac{1}{2}\mathbf{U}(\mathbf{U}^{\mathrm{T}}\mathbf{U} - \mathbf{V}^{\mathrm{T}}\mathbf{V}),$$

$$\nabla_{\mathbf{V}} F_\Omega(\mathbf{U},\mathbf{V}) = \nabla_{\mathbf{V}}\mathcal{L}_\Omega - \frac{1}{2}\mathbf{V}(\mathbf{U}^{\mathrm{T}}\mathbf{U} - \mathbf{V}^{\mathrm{T}}\mathbf{V}),$$

Where gradients of partial derivatives are calculated as

$$\nabla_{\mathbf{U}}\mathcal{L}_\Omega = \frac{1}{p}\mathbf{X}_L^{\mathrm{T}} P_\Omega(\mathbf{X}_L\mathbf{U}\mathbf{V}^{\mathrm{T}}\mathbf{X}_R^{\mathrm{T}} - \mathbf{L}^*)\mathbf{X}_R\mathbf{V},$$

$$\nabla_{\mathbf{V}}\mathcal{L}_\Omega = \frac{1}{p}\mathbf{X}_R^{\mathrm{T}} P_\Omega^{\mathrm{T}}(\mathbf{X}_L\mathbf{U}\mathbf{V}^{\mathrm{T}}\mathbf{X}_R^{\mathrm{T}} - \mathbf{L}^*)\mathbf{X}_L\mathbf{U}.$$

Note that for function $\mathcal{L}_\Omega$, its gradient of derivatives is

$$\nabla\mathcal{L}_\Omega = \frac{1}{p}P_\Omega(\mathbf{X}_L\mathbf{U}\mathbf{V}^{\mathrm{T}}\mathbf{X}_R^{\mathrm{T}} - \mathbf{L}^*).$$

Recall that we let **Z** = [**U**; **V**]. Correspondingly, we have

$$\tilde{F}_\Omega(\mathbf{Z}) = F_\Omega(\mathbf{U},\mathbf{V}).$$

For simplicity, we will ignore the notation of the observed index set, e.g., we will use $\tilde{F}(\mathbf{Z})$ to denote $\tilde{F}_\Omega(\mathbf{Z})$.

### A.1 Proof of Theorem 1

To prove Theorem 1, the following three lemmas are needed, where the detailed proof of Lemma 1 can be found in Zhang et al. (2018b) and that of Lemma 3 in Tu et al. (2015). We place the proof of Lemma 2 in appendix A.2.

**Lemma 1 (Local Convexity).** $\forall \mathbf{Z} = [\mathbf{U};\ \mathbf{V}] \in \mathbb{B}(\frac{\alpha\sqrt{\sigma_r}}{2})$, suppose $\|\mathbf{X}_L\mathbf{U}\|_{2,\infty} \leq 2\sqrt{\frac{\mu_0 r\sigma_1}{d_1}}$ and $\|\mathbf{X}_R\mathbf{V}\|_{2,\infty} \leq 2\sqrt{\frac{\mu_0 r\sigma_1}{d_2}}$; then there exist positive constants $c_1$ and $c_2$ such that if $|\boldsymbol{\Omega}| \geq c_1 \max\{\mu_0^2 r^2\kappa^2, \mu_0\mu_1 r\kappa n\}\log d$, then the following inequality holds with a probability of at least $1-\frac{c_2}{d}$ ($d = \max\{d_1, d_2\}$):

$$\langle \nabla\tilde{\mathbf{F}}(\mathbf{Z}), \mathbf{H}\rangle \geq \frac{1}{20}\|\mathbf{Z}\mathbf{Z}^{\mathrm{T}} - \mathbf{Z}^*\mathbf{Z}^{*\mathrm{T}}\|_F^2 + \frac{1}{4\sigma_1}\|\tilde{\mathbf{Z}}^*\tilde{\mathbf{Z}}^{*\mathrm{T}}\mathbf{Z}\|_F^2 + \frac{\sigma_r}{8}\|\mathbf{H}\|_F^2 - 40\|\mathbf{H}\|_F^4,$$

where $\tilde{\mathbf{Z}} = [\mathbf{U}^*;\ -\mathbf{V}^*]$.

**Lemma 2 (Local Smoothness).** Let $G_{\mathbf{U}} = \nabla_{\mathbf{U}}F_t(\mathbf{U},\mathbf{V}) - \mathbf{X}_L^{\mathrm{T}}\nabla\mathcal{L}_t(\tilde{\mathbf{L}})\mathbf{X}_R\mathbf{V} + \mathbf{X}_L^{\mathrm{T}}\nabla\mathcal{L}_\Omega(\tilde{\mathbf{L}})\mathbf{X}_R\mathbf{V}$, $G_{\mathbf{V}} = \nabla_{\mathbf{V}}F_t(\mathbf{U},\mathbf{V}) - \mathbf{X}_R^{\mathrm{T}}\nabla{\mathcal{L}_t}^{\mathrm{T}}(\tilde{\mathbf{L}})\mathbf{X}_L\mathbf{U} + \mathbf{X}_R^{\mathrm{T}}\nabla{\mathcal{L}_\Omega}^{\mathrm{T}}(\tilde{\mathbf{L}})\mathbf{X}_L\mathbf{U}$, and $G = [G_{\mathbf{U}};\ G_{\mathbf{V}}]$. $\forall \mathbf{Z} = [\mathbf{U};\ \mathbf{V}] \in \mathbb{B}(\frac{\sqrt{\sigma_r}}{4})$, suppose $\|\mathbf{X}_L\mathbf{U}\|_{2,\infty} \leq 2\sqrt{\frac{\mu_0 r\sigma_1}{d_1}}$, $\|\mathbf{X}_R\mathbf{V}\|_{2,\infty} \leq 2\sqrt{\frac{\mu_0 r\sigma_1}{d_2}}$; then, there exist positive constants $c_1$ and $c_2$ such that if $|\boldsymbol{\Omega}| \geq c_1\ \mu_0\mu_1 r\kappa n\log d$, the following inequality holds with a probability of at least $1-\frac{c_2}{d}$:

$$\begin{aligned} E\|G\|_F^2 \leq{}& (16r\sigma_1 + 4\sigma_1)\|\mathbf{Z}\mathbf{Z}^{\mathrm{T}} - \mathbf{Z}^*\mathbf{Z}^{*\mathrm{T}}\|_F^2 + \|\tilde{\mathbf{Z}}^*\tilde{\mathbf{Z}}^{*\mathrm{T}}\mathbf{Z}\|_F^2 + 4r\sigma_r^2\|\mathbf{H}\|_F^2 \\ &+ 128r\sigma_1\|\tilde{\mathbf{M}} - \mathbf{M}^*\|_F^2 + 16r\sigma_r^2\|\tilde{\mathbf{H}}\|_F^2, \end{aligned}$$

where $\tilde{\mathbf{Z}} = [\mathbf{U}^*;\ -\mathbf{V}^*]$, $\mathbf{M}^* = \mathbf{U}^*\mathbf{V}^{*\mathrm{T}}$, $\tilde{\mathbf{M}} = \tilde{\mathbf{U}}\tilde{\mathbf{V}}^{\mathrm{T}}$.

**Lemma 3.** $\forall \mathbf{Z}, \mathbf{Z}' \in \mathbb{R}^{(d_1+d_2)\times r}$, if $d(\mathbf{Z},\mathbf{Z}') \leq \frac{\|\mathbf{Z}'\|_2}{4}$, then the following inequality holds:

$$\|\mathbf{Z}\mathbf{Z}^{\mathrm{T}} - \mathbf{Z}'\mathbf{Z}'^{\mathrm{T}}\|_F \leq \frac{9}{4}\|\mathbf{Z}'\|_2 \times d(\mathbf{Z},\mathbf{Z}').$$

*Proof (Proof of Theorem 1).* By Definition 3, if we prove the inequality

$$E\|\tilde{\mathbf{H}}^s\|_F^2 \leq \rho E\|\tilde{\mathbf{H}}^{s-1}\|_F^2, \tag{8}$$

where $0 < \rho < 1$, then by induction, the proof of Theorem 1 is completed.

Now, we provide the detailed derivation of (8). Let

$$\begin{cases} G_{\mathbf{U}}^k = \nabla_{\mathbf{U}}F_{t_k}(\mathbf{U}^k,\mathbf{V}^k) - \mathbf{X}_L^{\mathrm{T}}\nabla\mathcal{L}_{t_k}(\tilde{\mathbf{L}})\mathbf{X}_R\mathbf{V}^k + \mathbf{X}_L^{\mathrm{T}}\nabla\mathcal{L}_\Omega(\tilde{\mathbf{L}})\mathbf{X}_R\mathbf{V}^k, \\ G_{\mathbf{V}}^k = \nabla_{\mathbf{V}}F_{t_k}(\mathbf{U}^k,\mathbf{V}^k) - \mathbf{X}_R^{\mathrm{T}}\nabla{\mathcal{L}_{t_k}}^{\mathrm{T}}(\tilde{\mathbf{L}})\mathbf{X}_L\mathbf{U}^k + \mathbf{X}_R^{\mathrm{T}}\nabla{\mathcal{L}_\Omega}^{\mathrm{T}}(\tilde{\mathbf{L}})\mathbf{X}_L\mathbf{U}^k. \end{cases}$$

According to the iteration of Algorithm 2, we have

$$\begin{cases} \mathbf{U}^{k+1} = P_{C_1}\left(\mathbf{U}^k - \tau G_{\mathbf{U}}^k\right), \\ \mathbf{V}^{k+1} = P_{C_2}\left(\mathbf{V}^k - \tau G_{\mathbf{V}}^k\right). \end{cases}$$

Since $t_k \in [n]$ and it is randomized from the uniform distribution, by the definition of the objective function, we have

$$\begin{cases} E(G_{\mathbf{U}}^k) = \nabla_{\mathbf{U}} F(\mathbf{U}^k, \mathbf{V}^k), \\ E(G_{\mathbf{V}}^k) = \nabla_{\mathbf{V}} F(\mathbf{U}^k, \mathbf{V}^k). \end{cases}$$

Note that $\mathbf{Z}_k = [\mathbf{U}^k; \mathbf{V}^k]$, so we denote $\mathbf{R}^k = \arg\min_{\mathbf{R}\in\mathbb{Q}_r} \|\mathbf{Z}^k - \mathbf{Z}^*\mathbf{R}\|_F$, $\mathbf{H}^k = \mathbf{Z}^k - \mathbf{Z}^*\mathbf{R}^k$, and $G^k = [G_{\mathbf{U}}^k; G_{\mathbf{V}}^k]$.

By induction, $\forall k \geq 0$, if $\mathbf{Z}^k \in \mathbb{B}(c_5\sqrt{\sigma_r})$, then

$$E\|\boldsymbol{H}^{k+1}\|_F^2 \leq E\|\boldsymbol{Z}^{k+1} - \boldsymbol{Z}^*\boldsymbol{R}^k\|_F^2 = E\|\boldsymbol{U}^{k+1} - \boldsymbol{U}^*\boldsymbol{R}^k\|_F^2 + E\|\boldsymbol{V}^{k+1} - \boldsymbol{V}^*\boldsymbol{R}^k\|_F^2 = E\|P_{C_1}\big(\boldsymbol{U}^k - \tau G_U^k\big) - \boldsymbol{U}^*\boldsymbol{R}^k\|_F^2 + E\|P_{C_2}\big(\boldsymbol{V}^k - \tau G_V^k\big) - \boldsymbol{V}^*\boldsymbol{R}^k\|_F^2 \leq E\|\boldsymbol{U}^k - \tau G_U^k - \boldsymbol{U}^*\boldsymbol{R}^k\|_F^2 + E\|\boldsymbol{V}^k - \tau G_V^k - \boldsymbol{V}^*\boldsymbol{R}^k\|_F^2 = E\|\boldsymbol{H}^k\|_F^2 - 2\tau\langle\nabla\tilde{F}(\boldsymbol{Z}^k), \boldsymbol{H}^k\rangle + \tau^2 E\|\boldsymbol{G}^k\|_F^2, \quad (9)$$

where the first inequality is obtained directly from the definition of $\mathbf{H}^{k+1}$ and $\mathbf{R}^k$, and the second inequality follows from the non-expansive property of the projection step $P_{c_i}$. The first equality holds because $\|[\mathbf{A}; \mathbf{B}]\|_F^2 = \|\mathbf{A}\|_F^2 + \|\mathbf{B}\|_F^2$, the second follows from the iteration of Algorithm 2, and the third follows from the definition of the inner product. By Lemma 1,

$$\langle\nabla\tilde{\mathbf{F}}(\mathbf{Z}^k), \mathbf{H}^k\rangle \geq \frac{1}{20}\|\mathbf{Z}^k\mathbf{Z}^{k^{\mathrm{T}}} - \mathbf{Z}^*\mathbf{Z}^{*^{\mathrm{T}}}\|_F^2 + \frac{1}{4\sigma_1}\|\tilde{\mathbf{Z}}^*\tilde{\mathbf{Z}}^{*^{\mathrm{T}}}\mathbf{Z}\|_F^2 + \frac{\sigma_r}{8}\|\mathbf{H}^k\|_F^2 - 40\|\mathbf{H}^k\|_F^4. \quad (10)$$

By Lemma 2,

$$E\|G^k\|_F^2 \leq (16r\sigma_1 + 4\sigma_1)\|\mathbf{Z}^k\mathbf{Z}^{k^{\mathrm{T}}} - \mathbf{Z}^*\mathbf{Z}^{*^{\mathrm{T}}}\|_F^2 + \|\tilde{\mathbf{Z}}^*\tilde{\mathbf{Z}}^{*^{\mathrm{T}}}\mathbf{Z}\|_F^2 + 4r\sigma_r^2\|\mathbf{H}^k\|_F^2 + 128r\sigma_1\|\widetilde{\mathbf{M}}^k - \mathbf{M}^*\|_F^2 + 16r\sigma_r^2\|\widetilde{\mathbf{H}}^k\|_F^2. \quad (11)$$

With (10) $\times\, \tau^2 +$ (11) $\times\, (-2\tau)$, we obtain

$$\begin{aligned} &-2\tau\langle\nabla\tilde{F}(\mathbf{Z}^k), \mathbf{H}^k\rangle + \tau^2 E\|\mathrm{G}^k\|_F^2 \\ &\leq \left(4\tau^2 r\sigma_r^2 - \frac{\tau\sigma_r}{8}\right)\|\mathbf{H}^k\|_F^2 - \frac{\tau\sigma_r}{8}\|\mathbf{H}^k\|_F^2 + 80\tau\|\mathbf{H}^k\|_F^4 \\ &+ \left[\tau^2(16r\sigma_1 + 4\sigma_1) - \frac{\tau}{10}\right]\|\mathbf{Z}^k\mathbf{Z}^{k^{\mathrm{T}}} - \mathbf{Z}^*\mathbf{Z}^{*^{\mathrm{T}}}\|_F^2 \\ &+ \left(\tau^2 - \frac{\tau}{2\sigma_1}\right)\|\tilde{\mathbf{Z}}^*\tilde{\mathbf{Z}}^{*^{\mathrm{T}}}\mathbf{Z}\|_F^2 + 128\tau^2 r\sigma_1\|\widetilde{\mathbf{M}}^k - \mathbf{M}^*\|_F^2 \\ &+ 16\tau^2 r\sigma_r^2\|\widetilde{\mathbf{H}}^k\|_F^2. \end{aligned}$$

By selecting $\tau \leq \min\{\frac{1}{40\sigma_1(4r+1)}, \frac{1}{32r\sigma_r}\}$ and $c_5 \leq \min\{\frac{1}{4}, \frac{1}{\sqrt{1280}}\}$ such that $\|\mathbf{H}^k\|_F^2 \leq c_5\sigma_r$, we have

$$\begin{aligned} &-2\tau\langle\nabla\tilde{F}(\mathbf{Z}^k), \mathbf{H}^k\rangle + \tau^2 E\|\mathrm{G}^k\|_F^2 \\ &\leq -\frac{\tau\sigma_r}{16}\|\mathbf{H}^k\|_F^2 + 128\tau^2 r\sigma_1\|\widetilde{\mathbf{M}}^k - \mathbf{M}^*\|_F^2 + 16\tau^2 r\sigma_r^2\|\widetilde{\mathbf{H}}^k\|_F^2. \end{aligned}$$

Note that here, $\widetilde{\mathbf{M}}^k = \widetilde{\mathbf{M}}^{s-1}$ and $\widetilde{\mathbf{H}}^k = \widetilde{\mathbf{H}}^{s-1}$. Since $\tilde{\mathbf{Z}}^{s-1} \in \mathbb{B}(\frac{\sqrt{\sigma_r}}{4})$, by Lemma3, we have

$$\|\widetilde{\mathbf{M}}^k - \mathbf{M}^*\|_F^2 \leq 6\sigma_1\|\widetilde{\mathbf{H}}^{s-1}\|_F^2;$$

thus,

$$-2\tau\langle\nabla\tilde{F}(\mathbf{Z}^k), \mathbf{H}^k\rangle + \tau^2 E\|\mathrm{G}^k\|_F^2 \leq -\frac{\tau\sigma_r}{16}\|\mathbf{H}^k\|_F^2 + (768\tau^2 \mathrm{r}\sigma_1^2 + 16\tau^2 \mathrm{r}\sigma_r^2)\|\widetilde{\mathbf{H}}^{s-1}\|_F^2. \quad (12)$$

Plugging (12) back into (9), we obtain

$$E\|\mathbf{H}^{k+1}\|_F^2 - E\|\mathbf{H}^k\|_F^2 \leq -\frac{\tau\sigma_r}{16}\|\mathbf{H}^k\|_F^2 + (768\tau^2 \mathrm{r}\sigma_1^2 + 16\tau^2 \mathrm{r}\sigma_r^2)\|\widetilde{\mathbf{H}}^{s-1}\|_F^2. \quad (13)$$

By first taking the summation of (13) over $k \in [m-1]$ and then taking the expectation with regard to every term on both sides, we finally obtain

$$E\|\mathbf{H}^m\|_F^2 - E\|\mathbf{H}^0\|_F^2 \le -\frac{\tau\sigma_r}{16}\sum_{k=0}^{m-1} E\|\mathbf{H}^k\|_F^2 + (768\tau^2 \mathrm{r}\sigma_1^2 + 16\tau^2 \mathrm{r}\sigma_r^2) mE\|\widetilde{\mathbf{H}}^{s-1}\|_F^2.$$

Note that since the relations

$$\|\mathbf{H}^m\|_F^2 \ge 0,$$
$$\mathbf{H}^0 = \widetilde{\mathbf{H}}^{s-1},$$
$$E\|\widetilde{\mathbf{H}}^s\|_F^2 = \frac{1}{m}\sum_{k=0}^{m-1} \|\mathbf{H}^k\|_F^2$$

exist, we then obtain

$$-E\|\mathbf{H}^0\|_F^2 = -E\|\widetilde{\mathbf{H}}^{s-1}\|_F^2 - \frac{\tau\sigma_r m}{16} E\|\widetilde{\mathbf{H}}^s\|_F^2 + (768\tau^2 \mathrm{r}\sigma_1^2 + 16\tau^2 \mathrm{r}\sigma_r^2) mE\|\widetilde{\mathbf{H}}^{s-1}\|_F^2.$$

Let $\rho = 16(\frac{1}{\tau\sigma_r m} + 768\tau \mathrm{r}\kappa\sigma_1 + 16\tau \mathrm{r}\sigma_r)$, then this inequality can be simplified as

$$E\|\widetilde{\mathbf{H}}^s\|_F^2 \le \rho E\|\widetilde{\mathbf{H}}^{s-1}\|_F^2.$$

Note that we can achieve $\rho \in (0,1)$ by selecting a small enough step size $\tau$ and a large enough $m = O(r)$. This completes our proof.

## A.2 Proof of Lemma 2

To prove Lemma 2, we need two corollaries from Zhang et al. (2018b). We refer readers to this paper for a detailed illustration of these two corollaries.

**Corollary 1.** $\forall \mathbf{Z} = [\mathbf{U};\ \mathbf{V}] \in \mathbb{B}(\frac{\sqrt{\sigma_r}}{4})$, suppose $\|\mathbf{X}_L\mathbf{U}\|_{2,\infty} \le 2\sqrt{\frac{\mu_0 r\sigma_1}{d_1}}$ and $\|\mathbf{X}_R\mathbf{V}\|_{2,\infty} \le 2\sqrt{\frac{\mu_0 r\sigma_1}{d_2}}$. Then, there exist positive constants $c_1$ and $c_2$ such that as long as $|\Omega| \ge c_1\ \mu_0\mu_1 r\kappa n\ \log d$, the following inequality holds with a probability of at least $1 - \frac{c_2}{d}$ ($d = \max\{d_1, d_2\}$):

$$\|\mathbf{X}_L^{\mathrm{T}}\nabla\mathcal{L}_t(\mathbf{L})\mathbf{X}_R\mathbf{V}^k\|_F^2 \le r(8\sigma_1\|\mathbf{M}-\mathbf{M}^*\|_F^2 + \sigma_r^2\|\mathbf{H}\|_F^2),$$

where $\mathbf{L} = \mathbf{X}_L\mathbf{U}\mathbf{V}^{\boldsymbol{T}}\mathbf{X}_R^{\mathrm{T}}$, $\mathbf{M} = \mathbf{U}\mathbf{V}^{\boldsymbol{T}}$, $\mathbf{H} = \mathbf{Z} - \mathbf{Z}^*\mathbf{R}_{opt}$, $\mathbf{R}_{opt} = arg\min_{\mathbf{R}\in\mathbb{Q}_r}\|\mathbf{Z} - \mathbf{Z}^*\mathbf{R}\|_F^2$.

**Corollary 2.** $\forall \mathbf{Z} = [\mathbf{U};\ \mathbf{V}] \in \mathbb{R}^{(d_1+d_2)\times r}$, the following inequality holds:

$$\|\mathbf{U}(\mathbf{U}^{\mathrm{T}}\mathbf{U} - \mathbf{V}^{\mathrm{T}}\mathbf{V})\|_F^2 + \|\mathbf{V}(\mathbf{U}^{\mathrm{T}}\mathbf{U} - \mathbf{V}^{\mathrm{T}}\mathbf{V})\|_F^2 \le 2\|\tilde{\mathbf{Z}}^*\tilde{\mathbf{Z}}^{*\mathrm{T}}\mathbf{Z}\|_F^2 + 8\sigma_1\|\mathbf{Z}\mathbf{Z}^{\mathrm{T}} - \mathbf{Z}^*\mathbf{Z}^{*\mathrm{T}}\|_F^2,$$

where $\tilde{\mathbf{Z}}^* = [\mathbf{U}^*;\ -\mathbf{V}^*]$.

*Proof (Proof of Lemma 2).* The main inspiration is to obtain upper bounds with regard to the norm for $G_{\mathbf{U}}$ and $G_{\mathbf{V}}$. First, we consider $G_{\mathbf{U}}$. By definition,

$$\begin{aligned}\|G_{\mathbf{U}}^k\|_F^2 &= \|\nabla_{\mathbf{U}} F_{t_k}(\mathbf{U}^k,\mathbf{V}^k) - \mathbf{X}_L^{\mathrm{T}}\nabla\mathcal{L}_{t_k}(\tilde{\mathbf{L}})\mathbf{X}_R\mathbf{V}^k + \mathbf{X}_L^{\mathrm{T}}\nabla\mathcal{L}_{\Omega}(\tilde{\mathbf{L}})\mathbf{X}_R\mathbf{V}^k\|_F^2 \\ &= \|\mathbf{X}_L^{\mathrm{T}}\nabla\mathcal{L}_{t_k}\left(\mathbf{X}_L\mathbf{U}^k\mathbf{V}^{k^{\boldsymbol{T}}}\mathbf{X}_R^{\mathrm{T}}\right)\mathbf{X}_R\mathbf{V}^k - \mathbf{X}_L^{\mathrm{T}}\nabla\mathcal{L}_{t_k}(\tilde{\mathbf{L}})\mathbf{X}_R\mathbf{V}^k \\ &\quad + \mathbf{X}_L^{\mathrm{T}}\nabla\mathcal{L}_{\Omega}(\tilde{\mathbf{L}})\mathbf{X}_R\mathbf{V}^k + \frac{1}{2}\mathbf{U}(\mathbf{U}^{\mathrm{T}}\mathbf{U} - \mathbf{V}^{\mathrm{T}}\mathbf{V})\|_F^2 \\ &\le 2\|\mathbf{X}_L^{\mathrm{T}}\nabla\mathcal{L}_{t_k}\left(\mathbf{X}_L\mathbf{U}^k\mathbf{V}^{k^{\boldsymbol{T}}}\mathbf{X}_R^{\mathrm{T}}\right)\mathbf{X}_R\mathbf{V}^k - \mathbf{X}_L^{\mathrm{T}}\nabla\mathcal{L}_{t_k}(\tilde{\mathbf{L}})\mathbf{X}_R\mathbf{V}^k \\ &\quad + \mathbf{X}_L^{\mathrm{T}}\nabla\mathcal{L}_{\Omega}(\tilde{\mathbf{L}})\mathbf{X}_R\mathbf{V}^k\|_F^2 + \frac{1}{2}\|\mathbf{U}(\mathbf{U}^{\mathrm{T}}\mathbf{U} - \mathbf{V}^{\mathrm{T}}\mathbf{V})\|_F^2,\end{aligned}$$

where the inequality holds because $(\mathrm{a}+\mathrm{b})^2 \le 2\mathrm{a}^2 + 2\mathrm{b}^2$.
Denote

$$I_1 = \|\mathbf{X}_L^{\mathrm{T}}\nabla\mathcal{L}_{t_k}\left(\mathbf{X}_L\mathbf{U}^k\mathbf{V}^{k^{\boldsymbol{T}}}\mathbf{X}_R^{\mathrm{T}}\right)\mathbf{X}_R\mathbf{V}^k - \mathbf{X}_L^{\mathrm{T}}\nabla\mathcal{L}_{t_k}(\tilde{\mathbf{L}})\mathbf{X}_R\mathbf{V}^k + \mathbf{X}_L^{\mathrm{T}}\nabla\mathcal{L}_{\Omega}(\tilde{\mathbf{L}})\mathbf{X}_R\mathbf{V}^k\|_F^2,$$
$$I_{2_{\mathbf{U}}} = \|\mathbf{U}(\mathbf{U}^{\mathrm{T}}\mathbf{U} - \mathbf{V}^{\mathrm{T}}\mathbf{V})\|_F^2;$$

Then,

$$\begin{aligned}E(I_1) &\le 3E\|\mathbf{X}_L^{\mathrm{T}}\nabla\mathcal{L}_{t_k}\left(\mathbf{X}_L\mathbf{U}^k\mathbf{V}^{k^{\boldsymbol{T}}}\mathbf{X}_R^{\mathrm{T}}\right)\mathbf{X}_R\mathbf{V}^k\|_F^2 + 3E\|\mathbf{X}_L^{\mathrm{T}}\nabla\mathcal{L}_{t_k}(\tilde{\mathbf{L}})\mathbf{X}_R\mathbf{V}^k\|_F^2 \\ &\quad + 3E\|\mathbf{X}_L^{\mathrm{T}}\nabla\mathcal{L}_{\Omega}(\tilde{\mathbf{L}})\mathbf{X}_R\mathbf{V}^k\|_F^2 \\ &= \frac{3}{n}\sum_{t=1}^{n}\|\mathbf{X}_L^{\mathrm{T}}\nabla\mathcal{L}_{t_k}\left(\mathbf{X}_L\mathbf{U}^k\mathbf{V}^{k^{\boldsymbol{T}}}\mathbf{X}_R^{\mathrm{T}}\right)\mathbf{X}_R\mathbf{V}^k\|_F^2 \\ &\quad + \frac{3}{n}\sum_{t=1}^{n}\|\mathbf{X}_L^{\mathrm{T}}\nabla\mathcal{L}_{t_k}(\tilde{\mathbf{L}})\mathbf{X}_R\mathbf{V}^k\|_F^2 + 3E\|\mathbf{X}_L^{\mathrm{T}}\nabla\mathcal{L}_{\Omega}(\tilde{\mathbf{L}})\mathbf{X}_R\mathbf{V}^k\|_F^2 \\ &= \frac{3}{n}\|\mathbf{X}_L^{\mathrm{T}}\nabla\mathcal{L}_{t_k}\left(\mathbf{X}_L\mathbf{U}^k\mathbf{V}^{k^{\boldsymbol{T}}}\mathbf{X}_R^{\mathrm{T}}\right)\mathbf{X}_R\mathbf{V}^k\|_F^2 + \frac{3}{n}\|\mathbf{X}_L^{\mathrm{T}}\nabla\mathcal{L}_{t_k}(\tilde{\mathbf{L}})\mathbf{X}_R\mathbf{V}^k\|_F^2 \\ &\quad + 3E\|\mathbf{X}_L^{\mathrm{T}}\nabla\mathcal{L}_{\Omega}(\tilde{\mathbf{L}})\mathbf{X}_R\mathbf{V}^k\|_F^2,\end{aligned} \tag{14}$$

where the first inequality holds because $(\mathrm{a}+\mathrm{b}+\mathrm{c})^2 \le 3(\mathrm{a}^2+\mathrm{b}^2+\mathrm{c}^2)$, and the second equality holds because all subsets have no intersection.

By Corollary 1, we have

$$\|\mathbf{X}_L^{\mathrm{T}}\nabla\mathcal{L}_{t_k}\left(\mathbf{X}_L\mathbf{U}^k\mathbf{V}^{k^{\boldsymbol{T}}}\mathbf{X}_R^{\mathrm{T}}\right)\mathbf{X}_R\mathbf{V}^k\|_F^2 \le r(8\sigma_1\|\mathbf{M}-\mathbf{M}^*\|_F^2 + \sigma_r^2\|\mathbf{H}\|_F^2), \tag{15}$$
$$\|\mathbf{X}_L^{\mathrm{T}}\nabla\mathcal{L}_{t_k}(\tilde{\mathbf{L}})\mathbf{X}_R\mathbf{V}^k\|_F^2 \le r\left(8\sigma_1\|\widetilde{\mathbf{M}}-\mathbf{M}^*\|_F^2 + \sigma_r^2\|\widetilde{\mathbf{H}}\|_F^2\right). \tag{16}$$

Plugging (15) and (16) back into (14), we obtain

$$\begin{aligned}E(I_1) &\le E\|\mathbf{X}_L^{\mathrm{T}}\nabla\mathcal{L}_{t_k}\left(\mathbf{X}_L\mathbf{U}^k\mathbf{V}^{k^{\boldsymbol{T}}}\mathbf{X}_R^{\mathrm{T}}\right)\mathbf{X}_R\mathbf{V}^k\|_F^2 + 4E\|\mathbf{X}_L^{\mathrm{T}}\nabla\mathcal{L}_{t_k}(\tilde{\mathbf{L}})\mathbf{X}_R\mathbf{V}^k\|_F^2 \\ &\le r(8\sigma_1\|\mathbf{M}-\mathbf{M}^*\|_F^2 + \sigma_r^2\|\mathbf{H}\|_F^2) + 4r\left(8\sigma_1\|\widetilde{\mathbf{M}}-\mathbf{M}^*\|_F^2 + \sigma_r^2\|\widetilde{\mathbf{H}}\|_F^2\right) \\ &\le 4r\sigma_1\|\mathbf{Z}\mathbf{Z}^{\mathrm{T}} - \mathbf{Z}^*\mathbf{Z}^{*\mathrm{T}}\|_F^2 + r\sigma_r^2\|\mathbf{H}\|_F^2 + 32r\sigma_1\|\widetilde{\mathbf{M}}-\mathbf{M}^*\|_F^2 \\ &\quad + 4r\sigma_r^2\|\widetilde{\mathbf{H}}\|_F^2.\end{aligned}$$

Therefore, we have

$$E\|G_{\mathbf{U}}^k\|_F^2 \le 8r\sigma_1\|\mathbf{Z}\mathbf{Z}^{\mathrm{T}} - \mathbf{Z}^*\mathbf{Z}^{*\mathrm{T}}\|_F^2 + 2r\sigma_r^2\|\mathbf{H}\|_F^2 + 64r\sigma_1\|\widetilde{\mathbf{M}}-\mathbf{M}^*\|_F^2 + 8r\sigma_r^2\|\widetilde{\mathbf{H}}\|_F^2 + \frac{1}{2}I_{2_{\mathbf{U}}}. \tag{17}$$

Similarly, we have

$$E\|G_{\mathbf{V}}^k\|_F^2 \le 8r\sigma_1\|\mathbf{Z}\mathbf{Z}^{\mathrm{T}} - \mathbf{Z}^*\mathbf{Z}^{*\mathrm{T}}\|_F^2 + 2r\sigma_r^2\|\mathbf{H}\|_F^2 + 64r\sigma_1\|\widetilde{\mathbf{M}}-\mathbf{M}^*\|_F^2 + 8r\sigma_r^2\|\widetilde{\mathbf{H}}\|_F^2 + \frac{1}{2}I_{2_{\mathbf{V}}}. \tag{18}$$

By Corollary 2,

$$I_{2_{\mathbf{U}}} + I_{2_{\mathbf{V}}} \leq 2\|\tilde{\mathbf{Z}}^*\tilde{\mathbf{Z}}^{*^{\mathrm{T}}}\mathbf{Z}\|_F^2 + 8\sigma_1\|\mathbf{Z}\mathbf{Z}^{\mathrm{T}} - \mathbf{Z}^*\mathbf{Z}^{*^{\mathrm{T}}}\|_F^2. \tag{19}$$

Combining (17)- (19), we can obtain

$$\begin{aligned}
E\|G\|_F^2 &= E\|G_{\mathbf{U}}^k\|_F^2 + E\|G_{\mathbf{V}}^k\|_F^2 \\
&\leq 16r\sigma_1\|\mathbf{Z}\mathbf{Z}^{\mathrm{T}} - \mathbf{Z}^*\mathbf{Z}^{*^{\mathrm{T}}}\|_F^2 + 4r\sigma_r^2\|\mathbf{H}\|_F^2 + 12864r\sigma_1\|\widetilde{\mathbf{M}} - \mathbf{M}^*\|_F^2 \\
&+ 16r\sigma_r^2\|\widetilde{\mathbf{H}}\|_F^2 + \|\tilde{\mathbf{Z}}^*\tilde{\mathbf{Z}}^{*^{\mathrm{T}}}\mathbf{Z}\|_F^2 + 4\sigma_1\|\mathbf{Z}\mathbf{Z}^{\mathrm{T}} - \mathbf{Z}^*\mathbf{Z}^{*^{\mathrm{T}}}\|_F^2 \\
&= (16r\sigma_1 + 4\sigma_1)\|\mathbf{Z}\mathbf{Z}^{\mathrm{T}} - \mathbf{Z}^*\mathbf{Z}^{*^{\mathrm{T}}}\|_F^2 + \|\tilde{\mathbf{Z}}^*\tilde{\mathbf{Z}}^{*^{\mathrm{T}}}\mathbf{Z}\|_F^2 + 4r\sigma_r^2\|\mathbf{H}\|_F^2 \\
&+ 128r\sigma_1\|\widetilde{\mathbf{M}} - \mathbf{M}^*\|_F^2 + 16r\sigma_r^2\|\widetilde{\mathbf{H}}\|_F^2,
\end{aligned}$$

which completes the proof of Lemma 2.